\documentclass[]{spie}  

\usepackage{amssymb}
\usepackage{pifont}

\usepackage{adjustbox}
 
\usepackage{amsmath,amsfonts,amssymb}
\usepackage{subcaption}
\usepackage{graphicx}
\usepackage{cite} 
\usepackage{times}
\usepackage{epsfig}
\usepackage{amsmath}
\usepackage{nccmath}
\usepackage{amssymb}
\usepackage{mwe}
\usepackage{acro}
\usepackage{amssymb}
\usepackage{xcolor,colortbl}
\usepackage{tabularx}
\usepackage{relsize}
\usepackage{pifont}
\usepackage{booktabs} 
\usepackage{multirow}
\usepackage{multicol}
\usepackage{adjustbox}
\usepackage{float}
\usepackage{graphicx}
\usepackage{makecell}
\usepackage{tabu}
\usepackage[colorlinks=true, allcolors=blue]{hyperref}
\usepackage[capitalize]{cleveref}

\title{Glo-VLMs: Leveraging Vision–Language Models for Fine‑Grained Diseased Glomerulus Classification}

\author[a]{Zhenhao Guo}
\author[b]{Rachit Saluja}
\author[c]{Tianyuan Yao}
\author[c]{Quan Liu}
\author[c]{Yuankai Huo}
\author[d]{Benjamin Liechty}
\author[d]{David J. Pisapia}
\author[d]{Kenji Ikemura}
\author[b,d]{Mert R. Sabuncu}
\author[d,e]{Yihe Yang}
\author[c,d]{Ruining Deng}

\affil[a]{New York University, New York, NY 10012, USA}
\affil[b]{Cornell Tech, New York, NY 10044, USA}
\affil[c]{Vanderbilt University, Nashville, TN, 37235, USA}
\affil[d]{Weill Cornell Medicine, New York, NY 10065, USA}
\affil[e]{Northwell Health, New Hyde Park, NY 11040, USA}

\begin{document} 
\maketitle

\begin{abstract}
Vision–language models (VLMs) have shown considerable potential in digital pathology, yet their effectiveness remains limited for fine-grained, disease-specific classification tasks such as distinguishing between glomerular subtypes. The subtle morphological variations among these subtypes, combined with the difficulty of aligning visual patterns with precise clinical terminology, make automated diagnosis in renal pathology particularly challenging. In this work, we explore how large pretrained VLMs can be effectively adapted to perform fine-grained glomerular classification, even in scenarios where only a small number of labeled examples are available.

In this work, we introduce Glo-VLMs, a systematic framework designed to explore the adaptation of VLMs to fine-grained glomerular classification in data-constrained settings. Our approach leverages curated pathology images alongside clinical text prompts to facilitate joint image–text representation learning for nuanced renal pathology subtypes. By assessing various VLMs architectures and adaptation strategies under a few-shot learning paradigm, we explore how both the choice of method and the amount of labeled data impact model performance in clinically relevant scenarios. To ensure a fair comparison, we evaluate all models using standardized multi-class metrics, aiming to clarify the practical requirements and potential of large pretrained models for specialized clinical research applications. As a result, fine-tuning the VLMs achieved 0.7416 accuracy, 0.9045 macro-AUC, and 0.5277 F1-score with only 8 shots per class, demonstrating that even with highly limited supervision, foundation models can be effectively adapted for fine-grained medical image classification.

\end{abstract}


\section{Description of purpose}
\label{sec:intro}  
Vision–language models (VLMs) have shown remarkable promise in digital pathology, excelling in tasks such as stain classification~\cite{seyfioglu2024quilt,li2023llava,song2023artificial}, tissue‑type identification~\cite{li2023llava,song2023artificial,dai2024pa}, and broad lesion recognition~\cite{song2023artificial,dai2024pa}. Their ability to combine visual and textual cues learned from large‑scale multimodal pretraining has opened new possibilities for computational pathology. Yet, these strengths do not fully extend to fine‑grained, disease‑specific problems. In particular, distinguishing between subtypes of diseased glomeruli remains challenging~\cite{yang2022glomerular,lu2021improve,yu2025glo,yao2022self}. The subtle morphological differences between classes make it difficult for pretrained VLMs to reliably match visual patterns to precise clinical terminology, and the alignment between image features and disease‑specific language often falls short. As a result, their diagnostic value in specialized renal pathology tasks is still limited.

Our preliminary findings highlight this gap: despite their demonstrated success in a variety of domains, both general-domain and pathology-adapted VLMs exhibit limited performance when tasked with diseased glomerular classification in a zero-shot context. This observation underscores the importance of domain-specific adaptation to bridge the semantic divide between subtle histopathological features and their corresponding clinical labels. To address this challenge, we introduce Glo-VLMs that systematically evaluate few-shot supervised fine-tuning approaches, wherein large pretrained models are adapted using only a limited number of labeled examples per class. This strategy offers a pragmatic compromise, allowing us to bypass the substantial data demands of full retraining while still tailoring the models to the nuanced requirements of this fine-grained classification task.

In this study, we examine the adaptability of three representative vision–language model (VLM) backbones in fine-grained diseased glomerular classification under limited-data conditions: CLIP~\cite{radford2021learning}, a general-domain model; PLIP~\cite{huang2023visual}, a pathology-specific variant of CLIP; and CONCH~\cite{lu2024visual}, a pathology-focused foundation model. Our evaluation covers four adaptation strategies: full fine-tuning of all model parameters~\cite{srinivasan2024comparative}, the incorporation of lightweight adapter modules~\cite{gao2024clip}, low-rank adaptation (LoRA)\cite{hu2022lora} applied to attention mechanisms, and classifier tuning, where the pretrained backbone is frozen and only the final prediction head is updated. For the classifier head, we compare three architectures: a simple linear projection, a multi-layer perceptron (MLP), and an MLP with Batch Normalization applied after the hidden layer\cite{ioffe2015batch}. Rather than seeking a definitive ranking of approaches, our goal is to understand how backbone selection, adaptation method, and data availability interact to shape model performance in this challenging clinical context. Through a series of few-shot learning experiments, we assess how different VLMs respond to limited supervision, offering practical insights for adapting large pretrained models to specialized clinical research tasks. Remarkably, with only eight labeled examples per class, fine-tuned VLMs reached 0.7416 accuracy, 0.9045 macro-AUC, and 0.5277 F1-score, highlighting their potential to deliver strong performance in fine-grained medical image classification, even with minimal annotated data.

\begin{figure}[h]
\begin{center}
\includegraphics[width=0.7\textwidth]{{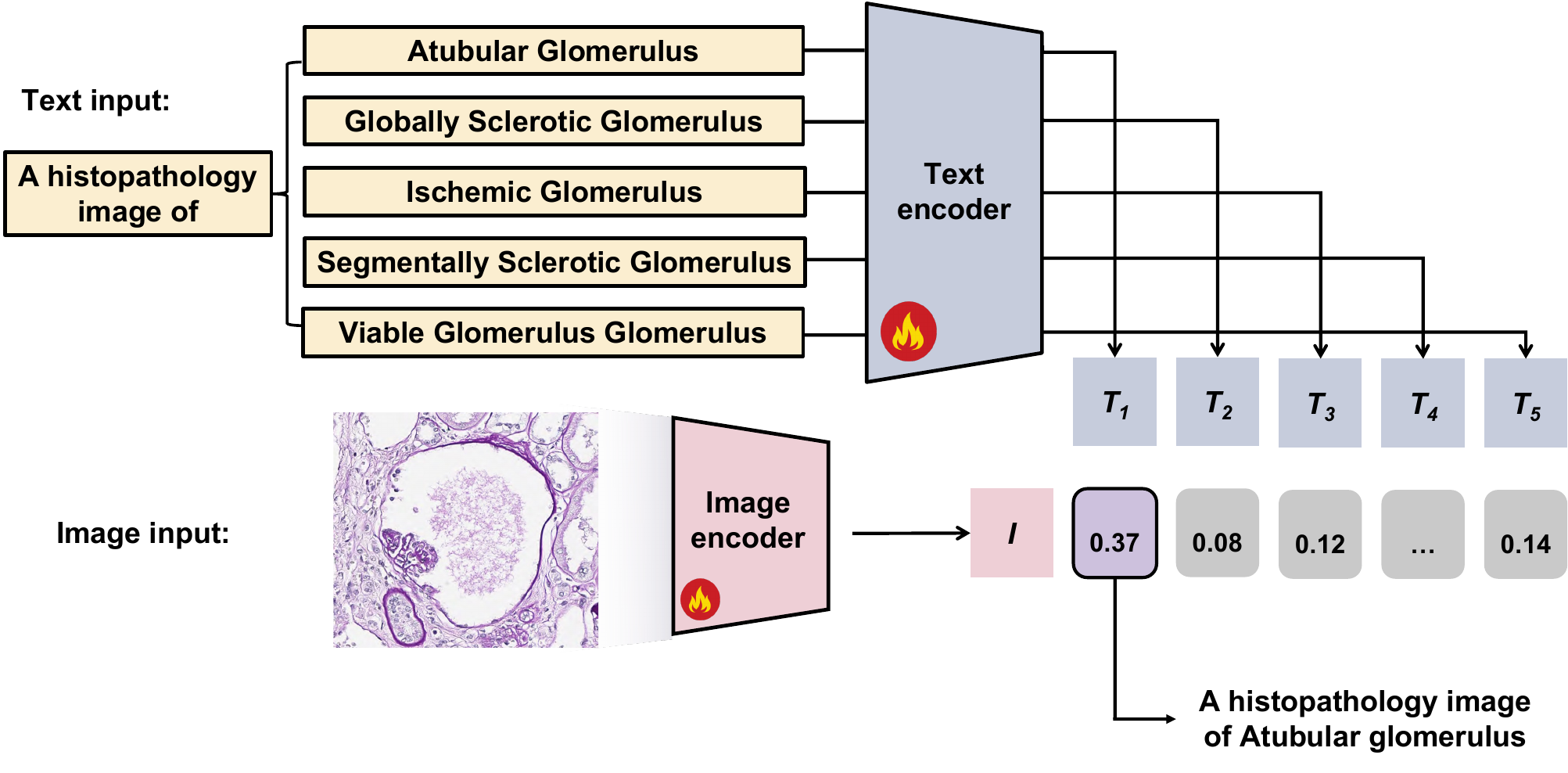}}
\end{center}
\caption{\textbf{Pipeline overview.} Whole-slide images-derived patches and class prompts are encoded by a pretrained VLMs. The backbone is fine-tuning with few-shots setting.}
\label{fig1:Pipeline Overview}
\end{figure}

\section{Method}

\subsection{Pipeline Overview}
To overcome the limited generalizability of existing methods for fine-grained pathological image classification, our approach centers on distinguishing among subtypes of diseased glomeruli within a low-data setting. Specifically, we investigate the adaptation of VLMs to a five-class glomerular pathology classification task, relying on only a small number of labeled examples per category. As depicted in Figure~\ref{fig1:Pipeline Overview}, the pipeline begins with pathological image patches paired with their corresponding class prompts (for example, ``A histopathology image of Atubular Glomerulus''). Both the images and textual prompts are encoded using a pretrained VLMs backbone, which is adapted to the task either through fine-tuning or the integration of lightweight modules. This process yields high-dimensional embeddings for both modalities. To generate predictions, we compute the cosine similarity between each image embedding and all class prompt embeddings, applying a softmax transformation to produce a probability distribution over the classes. The model then assigns each image to the class with the highest similarity score.

\subsection{Foundation VLMs Backbones}
To systematically assess the influence of model design, domain expertise, and modality alignment, we select three VLM backbones that differ in scale, architecture, and the extent of pathology-specific training. These are: a general-purpose foundation model (CLIP), a pathology-adapted variant (PLIP), and a pathology-native foundation model (CONCH).
\begin{itemize}
\item \textbf{CLIP}: A general-domain vision–language model trained on 400M image–text pairs using a contrastive objective. We use the openai/clip-vit-base-patch16 variant with a ViT~\cite{dosovitskiy2020vit} image encoder and Transformer-based text encoder, projecting both modalities into a shared embedding space for similarity-based matching.

\item \textbf{PLIP}: A pathology-adapted CLIP retrained on 200K pathology image–caption pairs to better align histopathology visuals with domain-specific terminology, yielding embeddings more sensitive to pathology-specific features.

\item \textbf{CONCH}: A pathology-native foundation model (ViT-B-16) based on the CoCa architecture~\cite{yu2022coca}, pretrained on 1.17M pathology image–caption pairs. It integrates contrastive and generative objectives with structured attention pooling to capture fine-grained multimodal relationships in high-resolution histology.
\end{itemize}

\begin{figure}[h]
\begin{center}
\includegraphics[width=0.7\textwidth]{{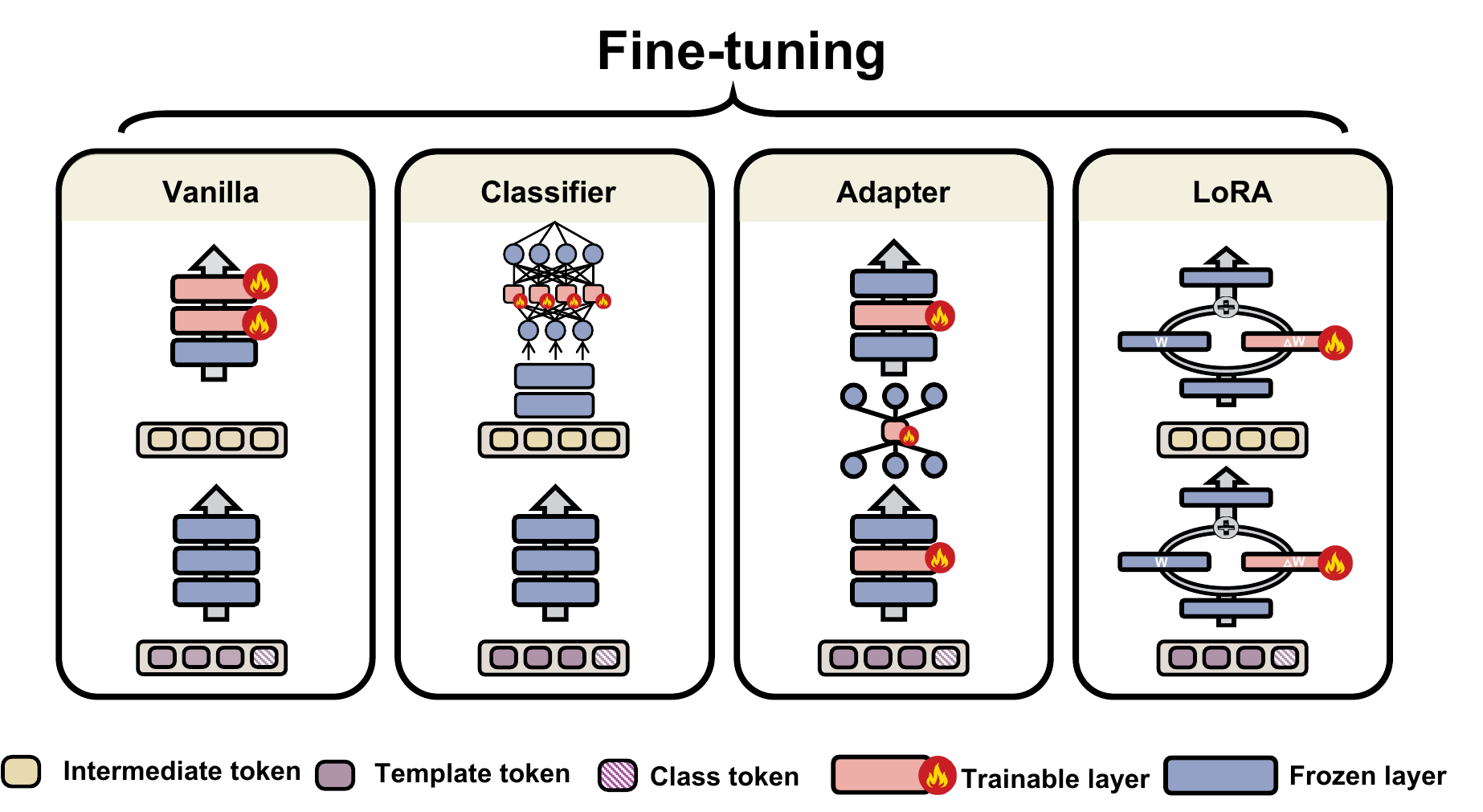}}
\end{center}
\caption{\textbf{Overview of fine-tuning strategies.} Schematic comparison of Vanilla fine-tuning, Classifier, Adapter, LoRA approaches.}
\label{fig2:Finetuning method}
\end{figure}
\subsection{Fine-tuning Strategies}

To adapt VLMs for fine-grained glomerular classification in few-shot settings, we examine four fine-tuning approaches, as illustrated in Figure~\ref{fig2:Finetuning method},spanning full- parameter updates to highly parameter-efficient methods~\cite{gao2024clip,srinivasan2024comparative}. Vanilla fine-tuning modifies both image and text encoders, while Adapter and LoRA insert lightweight modules into the encoder to adjust internal representations with minimal disruption. Classifier tuning updates only the prediction head, keeping the backbone fixed. All methods use a supervised contrastive objective, computing class probabilities via cosine similarity between image and text embeddings followed by softmax.

\begin{itemize}
\item \textbf{Vanilla Fine-tuning:}  
Unfreezes the top layers of image and text encoders for full adaptation, offering high capacity. By allowing these layers to adapt fully to the target task, vanilla fine-tuning provides high expressive capacity.
 Mitigated with a conservative learning rate, warm-up, and early stopping. 

\item \textbf{Low-Rank Adaptation (LoRA):}  
Adds trainable low-rank matrices to attention projections in top vision and text transformer blocks, learning in a low-dimensional subspace while preserving backbone knowledge. Rank ($r$) and scaling ($\alpha$) are tuned for capacity–efficiency balance.

\item \textbf{Adapter Tuning:}  
While preserving pretrained weights, inserts lightweight bottleneck adapters into selected vision encoder layers~\cite{houlsby2019parameter}, projecting down and back up before adding to the residual.

\item \textbf{Classifier Tuning:}  
Keeps the backbone frozen and trains only a classification head (linear or two-layer MLP, with/without batch normalization) to map features to class logits, followed by softmax.
\end{itemize}

\section{Data \& Experiments} 
\subsection{Data}
This study retrospectively analyzed eight nephrectomy cases processed at the Department of Pathology, Northwell Health, with specimens sourced from both Long Island Jewish Medical Center and Northshore University Hospital between July 2017 and October 2019. Whole-slide images (WSIs) were prepared from de-identified, non-cancerous sections of kidney parenchyma stained with periodic acid–Schiff (PAS) and scanned at 20$\times$ magnification using Leica GT450 RUO scanners. Glomeruli in each WSI were initially identified using QuPath, after which expert renal pathologists conducted meticulous manual annotations. Each glomerulus was outlined with a pixel-level instance mask and classified into one of five clinically relevant categories: global glomerulosclerosis (GGS), viable glomerulus, ischemic glomerulus, segmental glomerulosclerosis (SGS), or atubular glomerulus.

\subsection{Evaluation Metrics}
We evaluated all methods on the five-class diseased glomerulus classification task using a fixed train–validation split (seed 42) and standardized class prompts. Quantitative performance was assessed using overall accuracy, macro-averaged area under the ROC curve (AUC)~\cite{hanley1982meaning}, and macro-averaged F1 score, ensuring fair comparison through identical data augmentation and early stopping settings. For ROC analysis, we computed one-vs-rest curves and corresponding macro-AUC values for each backbone–method–shot configuration, with particular attention to sensitivity at low false positive rates.  Qualitatively, we examined the distribution of the predicted probability assigned to the true class across shots. For each backbone and fine-tuning method, box plots summarize the median, interquartile range, and variability in prediction confidence, providing insight into model calibration and stability under limited supervision.

\subsection{Data Preprocessing}
To standardize the model inputs, we begin by extracting image patches from the high-resolution WSIs. Individual glomeruli are first located using the provided annotation masks. For each identified glomerulus, we define a bounding box that is symmetrically expanded by 50 pixels on all sides to capture critical surrounding context. To ensure uniformity across samples, each bounding box is then reshaped to a square, yielding patches with consistent input dimensions. This approach guarantees that every patch contains one complete glomerulus, which not only preserves label purity but also prevents class ambiguity.

For the textual component, we generate class-specific prompts that serve as anchors for contrastive learning. Each of the five glomerular subtypes is represented by a standardized phrase in the form: ``A histopathology image of \{label\}.'' By keeping these prompts fixed across all experiments, we provide a stable and interpretable textual reference point to facilitate precise image–text alignment.

To ensure the validity of our few-shot experiments, we construct the training sets for 1, 2, 4, 8, 16, and 32 shots per class in a strictly nested manner, where each smaller-shot subset is fully contained within all larger-shot sets. This design guarantees that, for example, the 4-shot configuration contains the exact images from the 2-shot set plus two additional images per class, enabling a controlled comparison across different shot levels. The validation set is constructed from samples outside the 32-shot-per-class training pool, ensuring no data leakage and providing an unbiased evaluation of model performance.

\subsection{Experimental Setup and Configuration}
To ensure robust and stable model training, we conducted systematic hyperparameter optimization, exploring a range of learning rates and weight decay values. Model optimization was performed using the Adam optimizer, with a linear warm-up phase followed by a learning rate decay schedule to facilitate smooth convergence. To further enhance generalization and minimize overfitting, we consistently applied data augmentations from the Albumentations~\cite{info11020125} library and adopted early stopping, maintaining uniform settings for patience and minimum delta across all experiments.

For each fine-tuning strategy, we tailored the configuration as follows:
\begin{itemize}
\item \textbf{Vanilla Fine-tuning:} Unfreeze and optimize the top layers of both image and text encoders, refining high-level representations while retaining lower-layer knowledge.

\item \textbf{LoRA:} Insert trainable low-rank matrices into projection layers of top transformer blocks in both encoders, with rank ($r$) and scaling ($\alpha$) tuned via the PEFT framework~\cite{peft}. Injection points match those in vanilla fine-tuning for fair comparison.

\item \textbf{Adapter Tuning:} Add lightweight bottleneck adapters to the vision encoder after attention and within the MLP block~\cite{gao2024clip}, implemented via AdapterHub~\cite{pfeiffer2020AdapterHub} with varying bottleneck dimensions.

\item \textbf{Classifier Tuning:} Keep the backbone frozen and train a lightweight classification head (input 512, hidden 256, dropout 0.5) from scratch.
\end{itemize}

\begin{table}[ht]
\centering
\begin{adjustbox}{width=0.95\textwidth}
\begin{tabular}{ll|ccccccc|ccccccc|ccccccc}
\toprule
\multirow{2}{0.4in}{Model} & Method & \multicolumn{7}{c}{Accuracy (\%)} & \multicolumn{7}{c}{AUC (\%)} & \multicolumn{7}{c}{F1 (\%)} \\[0.3em]
\cmidrule{3-9}
\cmidrule{10-16}
\cmidrule{17-23}
& Shots & 0 & 1 & 2 & 4 & 8 & 16 & 32 & 0& 1 & 2 & 4 & 8 & 16 & 32 & 0 & 1 & 2 & 4 & 8 & 16 & 32 \\
\midrule

\midrule

\multirow{4}{*}{CLIP~\cite{radford2021learning}} &Vanilla~\cite{srinivasan2024comparative} & 0.2378 & 0.5536 & 0.3562 & 0.5841 & 0.4507 & 0.5603 & 0.7412 & 0.5445 & 0.6721 & 0.6898 & 0.7104 & 0.7897 & 0.8583 & 0.8949 & 0.1233 & 0.2679 & 0.2809 & 0.3547 & 0.3779 & 0.4441 & 0.5573 \\
&LoRA~\cite{hu2022lora} & 0.2378 & 0.2105 & 0.2129 & 0.3043 & 0.2355 & 0.3866 & 0.4088 & 0.5445  & 0.5644 & 0.5641 & 0.6110 & 0.6395 & 0.7191 & 0.7588 & 0.1233 & 0.1256 & 0.1268 & 0.1730 & 0.2095 & 0.2872 & 0.3529 \\
&Adapter~\cite{gao2024clip} & 0.2378 & 0.3486 & 0.3229 & 0.3803 & 0.5512 & 0.5366 & 0.6157 & 0.5445 & 0.6534 & 0.6220 & 0.6723 & 0.8062 & 0.8107 & 0.8462 & 0.1233 & 0.1728 & 0.2240 & 0.2004 & 0.3138 & 0.2626 & 0.4458 \\
&Classifier~\cite{ioffe2015batch}& 0.2378 & 0.4499 & 0.6007 & 0.6066 & 0.3399 & 0.6296 & 0.6913 & 0.5445 & 0.7118 & 0.7518 & 0.7776 & 0.7467 & 0.8626 & 0.8923 & 0.1233 & 0.2233 & 0.3321 & 0.3650 & 0.2972 & 0.4456 & 0.4991 \\
\midrule

\multirow{4}{*}{PLIP~\cite{huang2023visual}} &Vanilla~\cite{srinivasan2024comparative} & \textbf{0.5758} & \textbf{0.6272} & 0.6106 & 0.6047 & 0.6605 & 0.6894 & 0.7495 & 0.5068 & 0.7886 & 0.8089 & 0.8101 & 0.8406 & 0.8841 & 0.9072 & \textbf{0.1518} & 0.2793 & 0.3642 & 0.3966 & 0.4536 & 0.4720 & 0.5654 \\
&LoRA~\cite{hu2022lora} & \textbf{0.5758} & 0.5786 & 0.5837 & 0.5144 & 0.3633 & 0.2240 & 0.4214 & 0.5068 & 0.5300 & 0.5370 & 0.5822 & 0.5801 & 0.6566 & 0.7269 & \textbf{0.1518} & 0.1889 & 0.2107 & 0.2812 & 0.2562 & 0.2021 & 0.3276 \\
&Adapter~\cite{gao2024clip} & \textbf{0.5758} & 0.5718 & 0.6494 & 0.5216 & 0.2786 & 0.6173 & 0.7416 & 0.5068 & 0.7145 & 0.7480 & 0.6188 & 0.6120 & 0.7952 & 0.8582 & \textbf{0.1518} & 0.2549 & 0.3250 & 0.2770 & 0.2204 & 0.3634 & 0.5040 \\
&Classifier~\cite{ioffe2015batch} & \textbf{0.5758} & 0.4662 & 0.6332 & 0.5671 & 0.6130 & 0.6719 & 0.6886 & 0.5068 & 0.7035 & 0.8033 & 0.7930 & 0.8046 & 0.8678 & 0.8786 & \textbf{0.1518} & 0.2722 & 0.3883 & 0.3653 & 0.4259 & 0.4635 & 0.4889 \\
\midrule

\multirow{4}{*}{CONCH~\cite{lu2024visual}} &Vanilla~\cite{srinivasan2024comparative} & 0.0930 & 0.5394 & 0.5544 & 0.6743 & 0.6391 & 0.7329 & 0.7032 & \textbf{0.5913} & 0.7619 & 0.8062 & \textbf{0.8711} & 0.8810 & 0.9173 & 0.9227 & 0.0733 & 0.3242 & 0.3464 & 0.4374 & 0.4681 & 0.5159 & 0.5175 \\
&LoRA~\cite{hu2022lora} & 0.0930 & 0.0882 & 0.0882 & 0.0886 & 0.0890 & 0.1072 & 0.5251 & \textbf{0.5913} & 0.5819 & 0.5820 & 0.5825 & 0.6029 & 0.6378 & 0.7768 & 0.0733 & 0.0961 & 0.0959 & 0.0961 & 0.1100 & 0.1163 & 0.3337 \\
&Adapter~\cite{gao2024clip} & 0.0930 & 0.3763 & 0.3194 & 0.5627 & \textbf{0.7416} & \textbf{0.7665}& \textbf{0.7816} & \textbf{0.5913} & 0.6978 & 0.7971 & 0.8443 & 0.8922 & \textbf{0.9230} & \textbf{0.9294} & 0.0733 & 0.2682 & 0.2707 & 0.3932 & \textbf{0.5277} & \textbf{0.5630} & \textbf{0.5737} \\
&Classifier~\cite{ioffe2015batch} & 0.0930 & 0.6126 & \textbf{0.6842} & \textbf{0.6941} & 0.6965 & 0.7186 & 0.7353 & \textbf{0.5913} & \textbf{0.8339} & \textbf{0.8543} & 0.8659 & \textbf{0.9045} & 0.9144 & 0.9217 & 0.0733 & \textbf{0.3762} & \textbf{0.4369} & \textbf{0.4722} & 0.5093 & 0.5291 & 0.5420 \\
\midrule

\bottomrule
\end{tabular}
\end{adjustbox}
\caption{\textbf{Few-shot performance on five-class diseased glomerulus classification.} Accuracy, AUC, and F1 scores for three backbones (CLIP, PLIP, CONCH) under four fine-tuning strategies (Vanilla, LoRA, Adapter, Classifier) across shots ${0,1,2,4,8,16,32}$. ``Shots'' denotes the number of labeled images per class used for fine-tuning; $0$ indicates no fine-tuning.}
\label{tab:tabel1}
\end{table}

\section{Results}

\begin{figure}[h]
\begin{center}
\includegraphics[width=0.9\textwidth]{{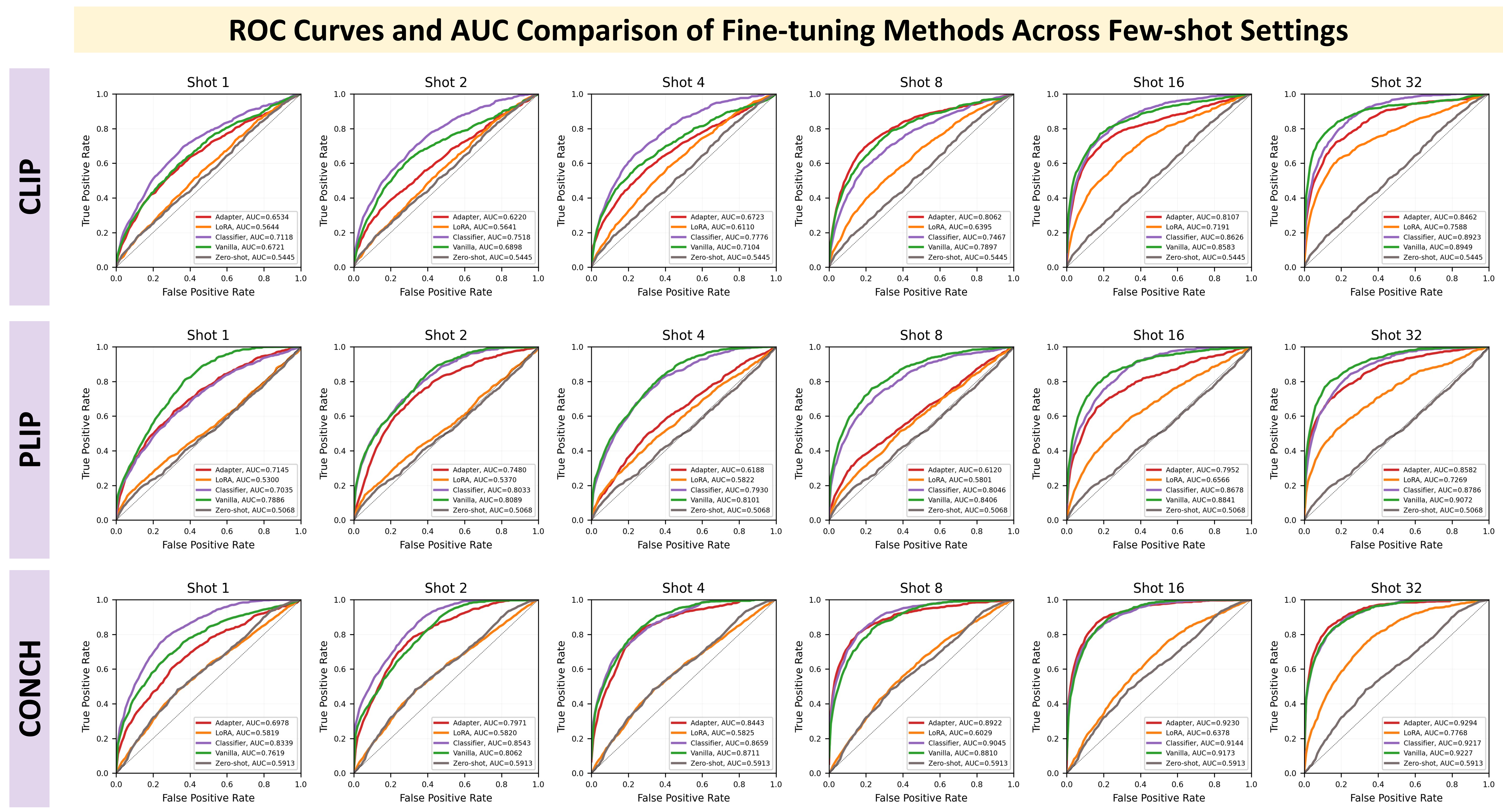}}
\end{center}
\caption{\textbf{ROC curves and AUC across few-shot settings.} Shot $0$ indicates no fine-tuning. }
\label{fig3: ROC}
\end{figure}

\begin{figure}
\begin{center}
\includegraphics[width=0.9\textwidth]{{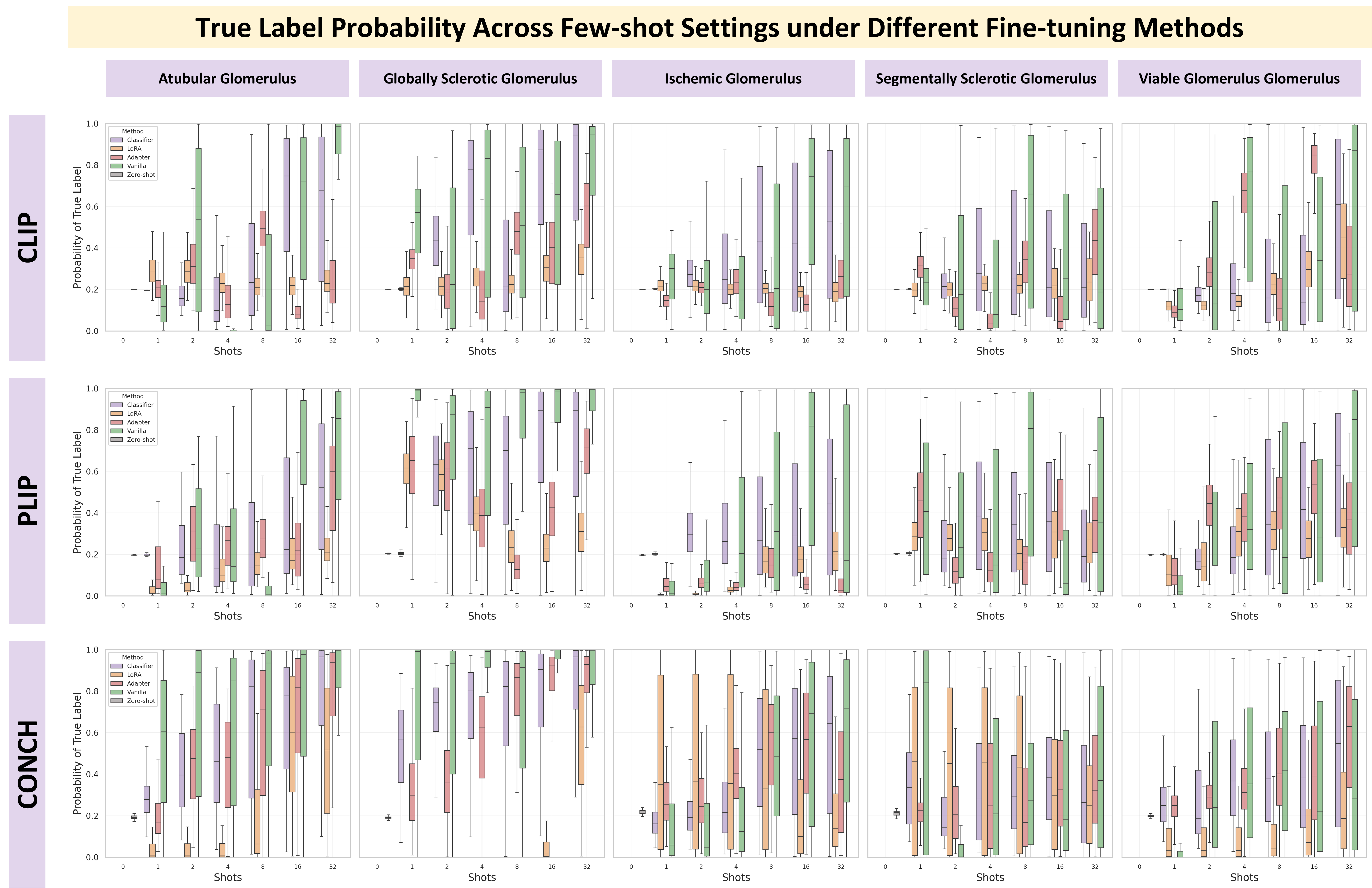}}
\end{center}
\caption{\textbf{True-label probability box plots across shots.} For each backbone and fine-tuning method, box plots show the predicted probability of the true class vs. shots.}
\label{fig4:boxplot}
\end{figure}

\subsection{Quantitative Results}

Table~\ref{tab:tabel1} and Figure~\ref{fig3: ROC} present few-shot classification results for different VLM backbones and fine-tuning strategies. Pathology-specific models (PLIP, CONCH) consistently outperform the general-domain CLIP, particularly in low-shot settings, underscoring the benefit of domain-aligned pretraining. As labeled samples increase, CLIP narrows the gap, especially at 16–32 shots, suggesting that few-shot adaptation can partially offset the lack of pathology-specific priors.
Among fine-tuning strategies, Classifier and Adapter tuning achieve the highest and most stable performance, while Vanilla fine-tuning shows steady improvement but remains slightly behind. LoRA trails in most scenarios, indicating a trade-off between parameter efficiency and adaptability. ROC curves confirm these patterns, with PLIP/CONCH paired with Classifier or Adapter tuning producing the strongest sensitivity–specificity trade-offs.

\subsection{Qualitative Results}

Qualitatively (Figure~\ref{fig4:boxplot}), PLIP shows the most stable and confident predictions, while CONCH demonstrates higher variability, especially in low-shot regimes. Adapter tuning maintains tight probability distributions with high medians in limited-data settings, whereas Vanilla fine-tuning catches up as supervision grows. Easier categories, such as Atubular Glomerulus and Global Glomerulosclerosis, achieve higher confidence and lower variability, while ischemic and segmental sclerotic glomeruli remain more challenging to distinguish.

\subsection{Discussion and Future Work}
Our results demonstrate that fine-tuned VLMs can achieve competitive performance in fine-grained diseased glomerulus classification, but important gaps remain. Future work will involve feature-level analyses, including embedding visualization and cross-modal similarity margin assessment, to determine whether improvements are driven by stronger positive clustering or by suppression of negatives. Model robustness should be validated through multiple seeds and repeated runs. The role of text encoder adaptation, particularly for pathology-specific terminology, also requires targeted ablation studies. Finally, evaluation on external datasets and exploration of semi-supervised or unsupervised adaptation may further improve generalizability and clinical applicability.

\section{New or Breakthrough Work to be Presented}
This study introduces the Glo-VLMs pipeline, a comprehensive framework for evaluating the adaptability of VLMs to fine-grained diseased glomerulus classification in few-shot learning scenarios. While prior research has largely emphasized zero-shot transfer or the adaptation of a single model, our work systematically examines multiple adaptation strategies across both general-domain and pathology-specialized backbones, enabling a nuanced assessment of their relative strengths under varying levels of supervision. By disentangling the interplay between backbone architecture, fine-tuning approach, and data availability, we provide empirical evidence on when domain-specific pretraining offers decisive advantages and when carefully tuned general-domain models can achieve comparable performance. These findings not only advance the methodological understanding of VLM adaptation in medical imaging but also deliver actionable guidelines for translating foundation-model capabilities into clinically relevant, data-efficient solutions.

\section{Conclusion}
In this study, we introduced the Glo-VLMs pipeline to systematically evaluate the adaptability of three vision–language model backbones for fine-grained diseased glomerulus classification under limited-data conditions. By assessing four adaptation strategies, we sought to elucidate how backbone architecture, fine-tuning approach, and data availability interact to shape model performance in a clinically demanding context. The results demonstrate that pathology-aware backbones consistently deliver the strongest performance in low-shot regimes, with parameter-efficient approaches such as Adapter tuning and Vanilla fine-tuning providing notable gains in efficiency without sacrificing accuracy. In contrast, the general-domain model exhibits substantial performance growth as labeled data increases, ultimately closing much of the gap with specialized models at higher shot counts. These findings highlight that optimal results depend on a deliberate alignment between backbone selection and fine-tuning strategy, underscoring that while pathology-specific pretraining offers a clear advantage in data-scarce scenarios, carefully designed adaptation can render general-domain models competitive in specialized medical image classification.

\section{ACKNOWLEDGMENTS} 
This research was supported by the WCM Radiology AIMI Fellowship and WCM CTSC 2026 Pilot Award. This research was also supported by NIH R01DK135597 and the KPMP Glue Grant.

\bibliography{main} 

\begin{thebibliography}{10}

\bibitem{seyfioglu2024quilt}
Seyfioglu, M.~S., Ikezogwo, W.~O., Ghezloo, F., Krishna, R., and Shapiro, L., ``Quilt-llava: Visual instruction tuning by extracting localized narratives from open-source histopathology videos,'' in [{\em Proceedings of the IEEE/CVF Conference on Computer Vision and Pattern Recognition}{\nolinebreak\hspace{0.1em}]},   13183--13192 (2024).

\bibitem{li2023llava}
Li, C., Wong, C., Zhang, S., Usuyama, N., Liu, H., Yang, J., Naumann, T., Poon, H., and Gao, J., ``Llava-med: Training a large language-and-vision assistant for biomedicine in one day,'' {\em Advances in Neural Information Processing Systems}~{\bf 36},  28541--28564 (2023).

\bibitem{song2023artificial}
Song, A.~H., Jaume, G., Williamson, D.~F., Lu, M.~Y., Vaidya, A., Miller, T.~R., and Mahmood, F., ``Artificial intelligence for digital and computational pathology,'' {\em Nature Reviews Bioengineering}~{\bf 1}(12),  930--949 (2023).

\bibitem{dai2024pa}
Dai, D., Zhang, Y., Xu, L., Yang, Q., Shen, X., Xia, S., and Wang, G., ``Pa-llava: A large language-vision assistant for human pathology image understanding,'' in [{\em 2024 IEEE International Conference on Bioinformatics and Biomedicine (BIBM)}{\nolinebreak\hspace{0.1em}]},   3138--3143, IEEE (2024).

\bibitem{yang2022glomerular}
Yang, C., Lee, C., Wang, H., Huang, S., Liang, P., Chen, J., Kuo, C., Tu, K., Yeh, C., and Chen, T., ``Glomerular disease classification and lesion identification by machine learning, biomed,'' (2022).

\bibitem{lu2021improve}
Lu, Y., Yang, H., Zhu, Z., Deng, R., Fogo, A.~B., and Huo, Y., ``Improve global glomerulosclerosis classification with imbalanced data using circlemix augmentation,'' {\em arXiv preprint arXiv:2101.07654}  (2021).

\bibitem{yu2025glo}
Yu, L., Yin, M., Deng, R., Liu, Q., Yao, T., Cui, C., Guo, J., Wang, Y., Wang, Y., Zhao, S., et~al., ``Glo-in-one-v2: holistic identification of glomerular cells, tissues, and lesions in human and mouse histopathology,'' {\em Journal of Medical Imaging}~{\bf 12}(6),  061406--061406 (2025).

\bibitem{yao2022self}
Yao, T., Lu, Y., Deng, R., Zhu, Z., Asad, Z., Yang, H., Wheless, L.~E., Fogo, A.~B., and Huo, Y., ``Self-supervised learning with large-scale web image mining for characterizing glomerular lesions,'' in [{\em Medical Imaging 2022: Digital and Computational Pathology}{\nolinebreak\hspace{0.1em}]},   {\bf 12039},  160--166, SPIE (2022).

\bibitem{radford2021learning}
Radford, A., Kim, J.~W., Hallacy, C., Ramesh, A., Goh, G., Agarwal, S., Sastry, G., Askell, A., Mishkin, P., Clark, J., et~al., ``Learning transferable visual models from natural language supervision,'' in [{\em International conference on machine learning}{\nolinebreak\hspace{0.1em}]},   8748--8763, PmLR (2021).

\bibitem{huang2023visual}
Huang, Z., Bianchi, F., Yuksekgonul, M., Montine, T.~J., and Zou, J., ``A visual--language foundation model for pathology image analysis using medical twitter,'' {\em Nature medicine}~{\bf 29}(9),  2307--2316 (2023).

\bibitem{lu2024visual}
Lu, M.~Y., Chen, B., Williamson, D.~F., Chen, R.~J., Liang, I., Ding, T., Jaume, G., Odintsov, I., Le, L.~P., Gerber, G., et~al., ``A visual-language foundation model for computational pathology,'' {\em Nature medicine}~{\bf 30}(3),  863--874 (2024).

\bibitem{srinivasan2024comparative}
Srinivasan, K. P.~V., Gumpena, P., Yattapu, M., and Brahmbhatt, V.~H., ``Comparative analysis of different efficient fine tuning methods of large language models (llms) in low-resource setting,'' {\em arXiv preprint arXiv:2405.13181}  (2024).

\bibitem{gao2024clip}
Gao, P., Geng, S., Zhang, R., Ma, T., Fang, R., Zhang, Y., Li, H., and Qiao, Y., ``Clip-adapter: Better vision-language models with feature adapters,'' {\em International Journal of Computer Vision}~{\bf 132}(2),  581--595 (2024).

\bibitem{hu2022lora}
Hu, E.~J., Shen, Y., Wallis, P., Allen-Zhu, Z., Li, Y., Wang, S., Wang, L., Chen, W., et~al., ``Lora: Low-rank adaptation of large language models.,'' {\em ICLR}~{\bf 1}(2),  3 (2022).

\bibitem{ioffe2015batch}
Ioffe, S. and Szegedy, C., ``Batch normalization: Accelerating deep network training by reducing internal covariate shift,'' in [{\em International conference on machine learning}{\nolinebreak\hspace{0.1em}]},   448--456, pmlr (2015).

\bibitem{dosovitskiy2020vit}
Dosovitskiy, A., Beyer, L., Kolesnikov, A., Weissenborn, D., Zhai, X., Unterthiner, T., Dehghani, M., Minderer, M., Heigold, G., Gelly, S., Uszkoreit, J., and Houlsby, N., ``An image is worth 16x16 words: Transformers for image recognition at scale,'' in [{\em International Conference on Learning Representations}{\nolinebreak\hspace{0.1em}]},  (2021).

\bibitem{yu2022coca}
Yu, J., Wang, Z., Vasudevan, V., Yeung, L., Seyedhosseini, M., and Wu, Y., ``Coca: Contrastive captioners are image-text foundation models,'' {\em arXiv preprint arXiv:2205.01917}  (2022).

\bibitem{houlsby2019parameter}
Houlsby, N., Giurgiu, A., Jastrzebski, S., Morrone, B., De~Laroussilhe, Q., Gesmundo, A., Attariyan, M., and Gelly, S., ``Parameter-efficient transfer learning for nlp,'' in [{\em International conference on machine learning}{\nolinebreak\hspace{0.1em}]},   2790--2799, PMLR (2019).

\bibitem{hanley1982meaning}
Hanley, J.~A. and McNeil, B.~J., ``The meaning and use of the area under a receiver operating characteristic (roc) curve.,'' {\em Radiology}~{\bf 143}(1),  29--36 (1982).

\bibitem{info11020125}
Buslaev, A., Iglovikov, V.~I., Khvedchenya, E., Parinov, A., Druzhinin, M., and Kalinin, A.~A., ``Albumentations: Fast and flexible image augmentations,'' {\em Information}~{\bf 11}(2) (2020).

\bibitem{peft}
Mangrulkar, S., Gugger, S., Debut, L., Belkada, Y., Paul, S., and Bossan, B., ``{PEFT}: State-of-the-art parameter-efficient fine-tuning methods.'' \url{https://github.com/huggingface/peft} (2022).

\bibitem{pfeiffer2020AdapterHub}
Pfeiffer, J., R{\"u}ckl{\'e}, A., Poth, C., Kamath, A., Vuli{\'c}, I., Ruder, S., Cho, K., and Gurevych, I., ``Adapterhub: A framework for adapting transformers,'' in [{\em Proceedings of the 2020 Conference on Empirical Methods in Natural Language Processing: System Demonstrations}{\nolinebreak\hspace{0.1em}]},   46--54 (2020).

\end{thebibliography}
\bibliographystyle{spiebib} 

\end{document}